\pdfoutput=1
\documentclass{article}
\usepackage{ijcai24}

\usepackage{times}
\usepackage{soul}
\usepackage{url}
\usepackage[hidelinks]{hyperref}
\usepackage[utf8]{inputenc}
\usepackage[small]{caption}
\usepackage{graphicx}
\usepackage{amsmath}
\usepackage{amsthm}
\usepackage{booktabs}
\usepackage{algorithm}
\usepackage{algorithmic}
\usepackage[switch]{lineno}
\usepackage{cleveref}
\usepackage{amssymb}
\usepackage{multirow}
\usepackage{multicol}
\usepackage{makecell}
\usepackage{diagbox}
\usepackage{xcolor}         
\usepackage{tabularx}

\usepackage{enumitem}

\title{A Survey on Graph Condensation}

\author{
Hongjia Xu$^1$
\and
Liangliang Zhang$^2$\and
Yao Ma$^{2}$\and
Sheng Zhou$^{1}$\thanks{Corresponding author}\and
Zhuonan Zheng$^{1}$\and
Bu Jiajun$^1$
\affiliations
$^1$Zhejiang University\\
$^2$Rensselaer Polytechnic Institute\\
\emails
\{xu\_hj, zhousheng\_zju, bjj, zhengzn\}@zju.edu.cn,
may13@rpi.edu,
liangliangz6v6@gmail.com
}

\begin{document}

\maketitle

\begin{abstract}
    Analytics on large-scale graphs have posed significant challenges to computational efficiency and resource requirements. 
    Recently,
    Graph condensation (GC) has emerged as a solution to address challenges arising from the escalating volume of graph data. 
    The motivation of GC is to reduce the scale of large graphs to smaller ones while preserving essential information for downstream tasks. 
    For a better understanding of GC and to distinguish it from other related topics, 
    we present a formal definition of GC and establish a taxonomy that systematically categorizes existing methods into three types based on its objective, and classify the formulations to generate the condensed graphs into two categories as modifying the original graphs or synthetic completely new ones.
    Moreover, our survey includes a comprehensive analysis of datasets and evaluation metrics in this field. 
    Finally, we conclude by addressing challenges and limitations, outlining future directions, and offering concise guidelines to inspire future research in this field.
\end{abstract}

\section{Introduction}
Graph data, representing relationships and interactions between entities, are ubiquitous in various domains including social networks, biological systems, and recommendation systems. Information and patterns in those scenarios have been modeled as nodes and edges, and there has been significant progress in the development of techniques for large-scale graph data mining and pattern recognition.

However, analyzing and processing large-scale graphs pose significant challenges to computational efficiency and resource requirements \cite{duan2022comprehensive}. Recently, the dataset distillation \cite{yu2023dataset} has attracted increasing attention and achieved success mainly in vision datasets.
Conventional dataset distillation relies on the idea that within categories defined by class labels, instances of the same class share similar key features, e.g. shape patterns in vision datasets. This implies the existence of 'prototypes' or 'clustering centers', and thus a significant amount of redundant information exists among instances belonging to the same category.
Similarly in graph datasets, e.g., in node classification tasks, the features and topology of nodes within the same class are similar, and there may be numerous repetitive and similar subgraph structures in the graph.
Thus a natural question arises: How can we effectively formulate small-scale graphs from large-scale graphs to facilitate various graph data mining tasks?

\textbf{Graph Condensation (GC)}, has been emerged for \textbf{distilling} large-scale graphs into smaller yet informative \textbf{new} ones.
By eliminating redundant information, GC makes the graph more manageable within the constraints of limited computation resources, thereby providing better support for graph data mining tasks and
applications such as Continual learning \cite{liu2023cat} and Network Architecture Search (NAS) \cite{gao2021graph}, etc. 
Moreover, take node classification as an example, the reason a node can be well classified is that GNNs have learned to capture the unique pattern of nodes to distinguish them from other nodes in different classes. Analogous to the attention heat map in the field of image classification (e.g. in \cite{lapuschkin2019unmasking}), the visualization of patterns that GNNs learned can be accomplished by GC.

While the concept aligns with vision dataset distillation, the specific challenges posed by the uniqueness of graph data motivate us to: (1) Address the lack of universal definitions and (2) Explore and synthesize the existing knowledge in this domain to a comprehensive survey.

\begin{figure*}[t]
  \centering
  \includegraphics[width=\linewidth]{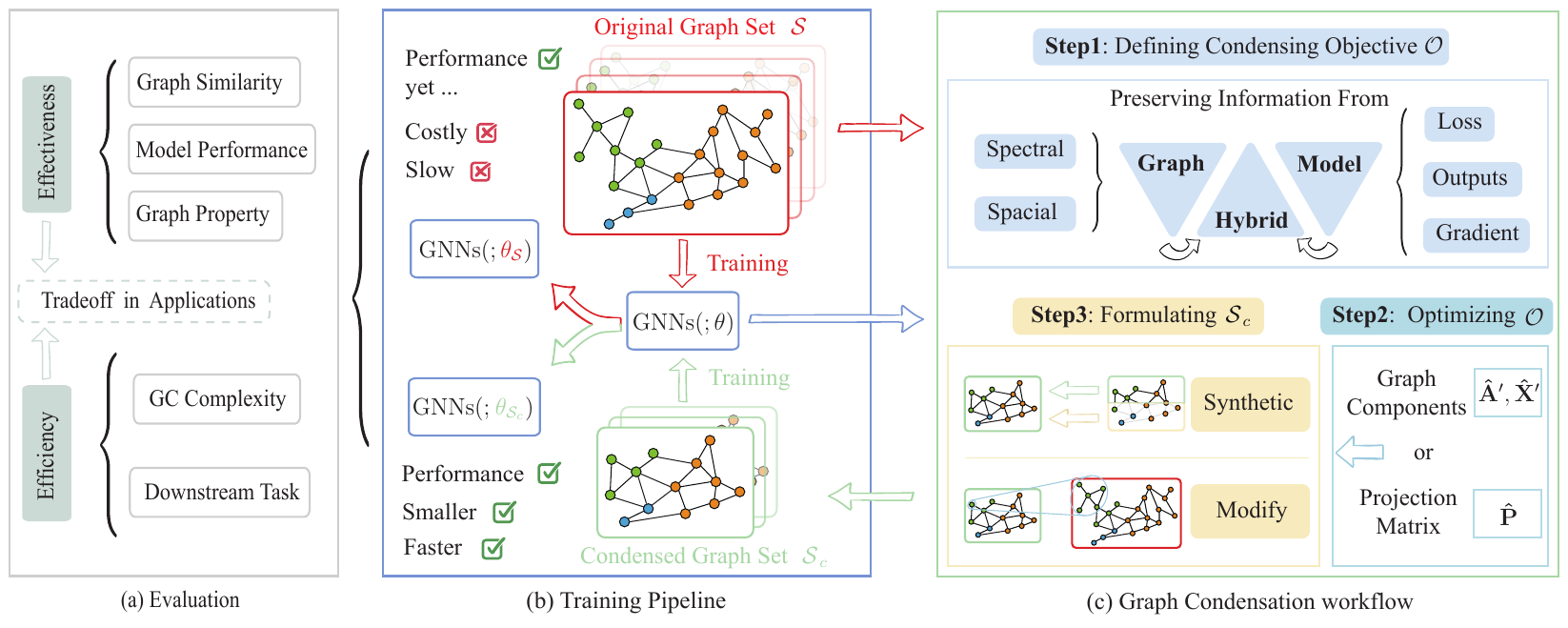} 
   \caption{Overview of GC. GNNs stand for any graph machine learning model with different architectures like GCN, GAT, ..., etc.}
   \label{fig-overview}
\vspace{-0.3cm}
\end{figure*}

\paragraph{Related topics.} 
The fundamental purpose of GC is to reduce the graph volume, however, there have been a few relevant topics that share a similar purpose:
\textbf{Graph Sampling} were designed to \textbf{select} subgraphs from original graphs, including
Core-set \cite{baker2020coresets} and subgraph mining \cite{nguyen2022subgraph} methods, etc. Nevertheless, sampling or pruning graph nodes or edges may cause massive information loss, resulting in performance collapse. See this paper \cite{littleballoffur} for a Python library on Graph Sampling; 
\textbf{Graph Reduction} \cite{loukas2019graph} intended to simplify the original graph to facilitate downstream tasks, yet mainly focus on simplifying the topology; 
\textbf{Graph Coarsen} \cite{chen2022graph} has been defined as an intermediate step of \textbf{Graph Pooling} \cite{grattarola2022understanding} in a recent survey \cite{liu2022graphpool}. 
Moreover, there are earlier surveys on a close subject, e.g., \textbf{Graph Summarization}\cite{liu2018graph,vcebiric2019summarizing}, and graph has been engaged as optional data modality under the definition of \textbf{Dataset Distillation} previously mentioned, as concurrent surveys did on the same topic \cite{gao2024graph,hashemi2024comprehensive}.

\paragraph{Scope of this paper.} 
To enhance the understanding of GC, we first propose a refined definition in \cref{definition}, which
clarifies the unique aspects of GC and highlights its divergence from vanilla dataset distillation. 
Under our definition, some methods of related topics are included in our discussion due to a shared motivation, i.e., facilitating graph data analytics by reducing the volume of graph data. Thus, the proposed taxonomy of GC is designed to encompass diverse techniques with various applications, and
our discussion extends to the consideration of various potential issues within GC.
Last but not least, we present various application scenarios and outline future directions.
Our contributions are as follows:
\begin{itemize}[leftmargin=1em]
    \item We present a formal definition of GC and systematically categorize existing methods into three types: graph-guided, model-guided, and hybrid. Through a detailed analysis, we classify the formulations of condensed graphs into modification and synthetic categories.
    \item We provide a summary of essential datasets, and conduct a thorough analysis of evaluation metrics and applications.
    \item We delve into the limitations and challenges of GC methods from a broader perspective, presenting future directions, and thus

    inspires future work in GC.
\end{itemize}

\section{Preliminary}
\subsection{Notations and Definition}
For any matrix, the symbol 
$\top,-,+$ represents the operations of transpose, inverse, and pseudoinverse, respectively.
Consider $\mathcal{S}=\{\mathcal{G}_1, \cdots, \mathcal{G}_{\eta}\}$ denote a dataset of $\eta$ graph(s), where $\eta\in \mathbb{N}$, $\mathcal{G}=\{\mathcal{V},\mathcal{E}, {\bf A}, {\bf X}\}$, $\mathcal{V}$ and $\mathcal{E}$ denotes the set of vertices (nodes) and edges, ${\bf A} \in \mathbb{R}^{N \times N}$ is the adjacency matrix, and ${\bf X} \in \mathbb{R}^{N \times d}$ represents the feature matrix. ${\bf L}_{\mathcal{G}}$ is the corresponding Laplacian matrix of $\mathcal{G}$.
$\mathcal{Y}\in \{ 1,\cdots, C \}^M$ denotes the set of labels of the nodes or edges if $\eta=1$, or the label of graphs if $\eta>1$, $[\mathcal{Y}] \in \mathbb{N}^{M \times 1}$ be its vector form. 
The downstream tasks are specifically defined by $M=N$ for node classification, $M=N^2$ for link prediction, and $M=\eta$ for graph classification. 
$\operatorname{GNN}(;{\theta}_{\mathcal{S}}): \mathcal{S}\rightarrow\mathcal{Y}$ denotes the GNN parameterized with ${\theta}_{\mathcal{S}}$ was trained on dataset $\{\mathcal{S},\mathcal{Y}\}$.

\subsection{Problem Defination}
\label{definition}

Denote $\mathcal{S}_c=\{{\mathcal{G}_c}_1, \cdots, {\mathcal{G}_c}_{\mu}\}$ 
a smaller dataset of $\mu$ graph(s), $\mu \leq \eta$, and $\mathcal{G}_c=\{\mathcal{V}^{\prime},\mathcal{E}^{\prime},{\bf A}^{\prime}, {\bf X}^{\prime}\}$ where the first dimension of $\mathcal{V}^{\prime}, {\bf A}^{\prime}$, and ${\bf X}^{\prime}$ are $N^{\prime}\left(N^{\prime} \ll N\right)$, $\mathcal{Y}^{\prime}\in \{ 1,\cdots, C \}^{M^{\prime}}$ be the labels of $\mathcal{G}_c$.
The conventional definition of GC \cite{gao2024graph} is to directly borrow the definition of the dataset distillation problem, considering graphs as input and output:
\begin{equation}
\mathcal{S}_c=\arg \min _{\mathcal{S}_c} \mathcal{L}_{\text {cond }}(GNN(\{\mathcal{S}_c, \mathcal{Y}^{\prime}\}, \{\mathcal{S},\mathcal{Y}\}; \theta))
\end{equation}
where $\mathcal{L}_{\text {cond }}$ is the optimization objective of condensation.

In contrast, we define Graph Condensation ($GC():\mathcal{S}\rightarrow \mathcal{S}_c$), a class of methods aimed to
scaling large-scale graphs to 
smaller yet informative \textbf{NEW} ones. We call:
\begin{equation}
\begin{aligned}
        \mathcal{G}_c &\text{ as the \textbf{Condensed} graph only if }\mathcal{V}^{\prime} \nsubseteq \mathcal{V}.\\
        \mathcal{S}_c &\text{ as the \textbf{Condensed} graphs only if } \mathcal{S}_c^{\prime} \nsubseteq \mathcal{S}. 
\end{aligned}
\end{equation}

In this paper, we will focus on condensing a single graph, while methods for multiple will be briefly introduced.
To ensure the condensed graphs are informative, 
their formulations are parameterized through an optimization process:
\begin{equation}
        \mathcal{G}_c = f(\mathcal{G}; {\bf {W}}),
        {\bf {W}} = \arg \min _{{\bf W}} \mathcal{O} (\phi({\mathcal{G}_c},\mathcal{Y}^{\prime}), \phi({\mathcal{G},\mathcal{Y}}))
\end{equation}
\begin{itemize}
    \item Condensation Objective $\mathcal{O}$ describes the loss of graph information, which is quantified by function $\phi$;
    \item ${\bf W}$ is optimized by minimizing the objective $\mathcal{O}$.
    \item The Formulation function $f()$ describes how we formulate the condense graphs, which is parameterized by ${\bf {W}}$;
\end{itemize}

The three steps for formulating condensed graphs correspond to the three steps in the GC workflow as illustrated in \cref{fig-overview}. Under our definition, specifying what information to preserve, denoted as $\phi$, is crucial as the primary motivation is to reduce the scale of graph data while preserving sufficient information. 
The details of the optimization objectives be seen in \cref{objectivesec}, and the formulations will be presented in \cref{approachsec}. We categorize current methods by the taxonomy of objectives and present the formulations of each in \cref{taxonomy tabel}.

\section{Condensation Objective}
\label{objectivesec}

We categorize the condensing objectives into three types: preserving certain properties of the graph (\textbf{graph guided}), retaining the GNNs' capabilities for downstream tasks (\textbf{model guided}), or simultaneously accomplishing both (\textbf{hybrid}).

\subsection{Graph Property Guided Methods}

These kinds of objectives can be formulated as:
Get similar \& smaller graph $\mathcal{G}_c$ from original graph $\mathcal{G}$, and the key is defining what are \textbf{graph properties} and how to \textbf{evaluate the similarity} between two graphs based on these properties. In this case, $\phi$ be graph property extractor, and $\mathcal{O}$ be the corresponding similarity/distance function.
We further categorize these objectives into two categories: \textbf{Spectral} and \textbf{Spacial}, by the domain of extracted information. Specifically:

\paragraph{Spectral Property Guided Methods.}
The Spectral GNNs \cite{chen2023bridging} are defined by operators in the Spectral domain.
Similarly, we define the \underline{\textsl{Spectral properties}} of a graph by requiring the graph Laplacian for calculations, i.e., $\phi(\mathcal{G})=\phi({\bf L}_{\mathcal{G}})$. In this case, the objective becomes minimizing the \textbf{D}istance of two graphs in the \textbf{Spe}ctral domain (\textbf{DSpe}), i.e., $\mathcal{O}=DSpe(\phi({\bf L}_{\mathcal{G}_c}),\phi({\bf L}_{\mathcal{G}}))$.

The direct use of Laplacian eigenvalue and eigenvectors can be seen in this domain, as well as the Laplacian Energy Distribution (LED) used in \cite{yang2023does}, which was derived from graph Laplacian.
However, the difference in graph scale before and after condensation may need cross-dimension metrics, such as having a distinct number of eigenvalues and eigenvectors. Specifically, \cite{loukas2019graph,huang2021scaling} calculate the differences of smallest $k$ eigenvalues and corresponding eigenvectors between two graphs' Laplacian;
\cite{liu2023graph} take $k$ eigenvectors with the smallest eigenvalues to map node features, and minimize the distances of class centers in the spectral space. This is because the smaller the eigenvalue is, the more informative it and its corresponding eigenvector are on graph laplacian\cite{das2004laplacian}. \cite{jin2020graph} Further propose graph lifting to rescale the small graph to a large one, and thus comparing all eigenvalues. Another issue is the efficiency concern, and all methods mentioned above have their Laplacian approximation design, which will not be discussed here.

Despite direct optimization of spectral properties, \cite{jin2020graph,deng2020graphzoom} identify similar nodes to \underline{\textsl{Aggregate}} in spectral embedding space, which can be seen as indirectly minimizing the spectral similarity of two graphs. Notably, minimizing \underline{\textsl{Graph similarity}} metrics based on spectral-GNNs, e.g. in \cite{liu2023cat}, is also considered as spectral property guided objectives with $\phi(\mathcal{G})=GNN(\mathcal{G})$, and similarly, the use of spatial-GNNs for similarity metrics applies to the subsequent definition of Spacial ones. The GNNs we discussed here were not trained on downstream tasks and only served as information extractors.

\paragraph{Spacial Property Guided Methods.}
The spatial domain of a graph is the original topology and node features, i.e., $\phi(\mathcal{G})=\phi(\{{\bf A},{\bf X}\})$. Objective becomes minimizing the \textbf{D}istance of two graphs in the \textbf{Spa}cial domain (\textbf{DSpa}), i.e., $\mathcal{O}=DSpa(\phi(\{{\bf A^{\prime}},{\bf X^{\prime}}\}),\phi(\{{\bf A},{\bf X}\}))$.
Specifically:

\underline{\textsl{Graph Statistic}} properties ($\phi(\mathcal{G})=Statistics({\bf A})$) like graph density, average degree and degree variance employed in \cite{xu2023better}, and feature homophily ($\phi(\mathcal{G})=Statistics(\{{\bf A},{\bf X}\})$) in \cite{kumar2023featured}. Moreover, Structural Equivalence Matching (SEM) and Normalized Heavy Edge Matching (NHEM) used in \cite{liang2021mile} can identify \underline{\textsl{Topologically Redundant}} nodes. Selecting nodes by \underline{\textsl{Ranking}} them through score functions can be effective in well-designed scenarios \cite{wei2023clnode}, while simple selection without aggregation will not be further discussed for not fitting our definition.

A special class of \underline{\textsl{Reconstruction}} like objectives is also considered this category of objectives because a successful reconstruction can be seen as a successful preservation of graph information. In this case, $\mathcal{O}=\mathcal{L}_{reconstruct}(\mathcal{G}_c,\mathcal{G})$. For example, metrics on reconstructed node features are used in \cite{ma2021unsupervised,kumar2023featured}, and metrics on reconstructing the whole graph in \cite{gao2023graph}.

\subsection{Model Capability Guided Methods}

Since the ultimate objective is to achieve comparable performance via training models (include but not limited to GNNs) on smaller graphs $\mathcal{S}_c$, the \textbf{models trained on the original graphs} $\mathcal{S}$ can be useful. 
By expecting GNNs to achieve \textbf{comparable results} as those trained on the original dataset $\mathcal{S}$ through training on the condensed ones $\mathcal{S}_c$, we write:

\begin{equation}
\label{equa2}
    \begin{aligned}
        &\min _{\mathcal{G}_c} \mathcal{L}\left(\mathrm{Model}(S;{\boldsymbol{\theta}_{\mathcal{G}_c}}), \mathcal{Y}\right) \\ 
        \text { s.t } \quad &\boldsymbol{\theta}_{\mathcal{G}_c}=\underset{\boldsymbol{\theta}}{\arg \min } \mathcal{L}\left(\mathrm{Model}\left(\mathcal{G}_c; {\boldsymbol{\theta}}\right), \mathcal{Y}^{\prime}\right),
    \end{aligned}
\end{equation}
where $\mathcal{L}$ is the task-specific loss (performance) function.

In this case, $\phi(\{\mathcal{G},\mathcal{Y}\})=\phi(\mathrm{Model}(;{\boldsymbol{\theta}_{\mathcal{G}}}))$, and $\mathcal{O}=D(\phi(\mathrm{Model}(;{\boldsymbol{\theta}_{\mathcal{G}_c}})),\phi(\mathrm{Model}(;{\boldsymbol{\theta}_{\mathcal{G}}})))$, $D$ is a distance function.
We categorize all objectives that utilize such trained model as input as model-guided objectives. Specifically:

\paragraph{Gradient.} Starting from \cite{jin2021graph}, training trajectory matching, i.e., aligning the \underline{\textsl{Gradient}} of models' parameters, has become one of the mainstream objectives in this category as \cite{jin2022condensing,li2023attend,yang2023does,zheng2023structure,gao2023graph,gao2023multiple} did.
Despite the specific model and tasks, these methods treat condensed graphs as optimization parameters to simulate steps of the models' training between the original graph and the condensed graph, and \cite{li2023attend,mao2023gcare} further introduce adversarial training for optimization. 
By doing so, the models trained on the condensed graph align well with the original models, maximizing the preservation of models' performances on specific tasks.

\paragraph{Loss Value.} Instead of applying GNNs as a black-box model, \cite{xu2023kernel,wang2023fast} aim to obtain the exact solution of the classification model from data. Specifically, kernel ridge regression was selected as the classifier model, and the \underline{\textsl{Loss value}} evaluating the performance of the condensed model on original data was optimized.

\paragraph{Embedding and Logits.} The outputs of an instance through trained network typically integrate crucial information for downstream tasks, and is thus considered informative. Specifically, matching models' output \underline{\textsl{Embeddings}} of training instances is used in \cite{liu2022graph,liu2023cat,dickens2023graph}, and \underline{\textsl{Predicted Logits}} based uncertainty metric was used in \cite{liu2023graph,xu2023better}.

\subsection{Hybrid Methods} 

It is worth mentioning that the aforementioned two types of objectives are not mutually conflicting. 
Therefore, the third category named hybrid methods combines both the graph properties and model capabilities as guidance during condensation simultaneously. 
There are methods such as \cite{yang2023does,gao2023graph,liu2023graph,xu2023better} that optimize the condensed graph from both graph-guided and model-guided objectives.
Specifically, \cite{yang2023does} simultaneously match the train trajectory between two models and the Laplacian Energy Distribution between two graphs, \cite{gao2023graph} optimize the training trajectory loss and reconstruction loss together, \cite{liu2022graph} perform eigenbasis and training trajectory matching in the same time and \cite{xu2023better} take the model predicted uncertainty and empirically verified useful graph properties to rank graph training instances for selection.

\subsection{Comparison of Objectives}
Three types of objectives, namely graph-guided, model-guided, and hybrid, each with its advantages and drawbacks: To produce 'similar' condensed graphs, \textbf{graph-guided} objectives focus on preserving the properties of the original graph. This is suitable for applications that require retaining the patterns from original graphs. 
However, they are not guided by downstream tasks and hence may not be the optimal solution.
On the other hand, the \textbf{model-guided} objectives aim to maintain the performance of the model by optimizing the condensed graph. These methods are driven by motivation-oriented optimization and thus perform exceptionally well in predefined scenarios. However, it may result in overfitting, reducing the adaptability of condensed graphs to other models or tasks.
\textbf{Hybrid} methods combine the advantages of both graph-guided and model-guided approaches, intending to retain model performance while preserving graph properties for scenarios that value both graph property and model performance. However, balancing between the two objectives as well as optimizing them can be challenging.

In conclusion, the choice of the appropriate objective depends on the specific requirements of the application. Graph-guided is more suitable for tasks emphasizing graph structure, model-guided applies to scenarios emphasizing model performance, and the hybrid method seeks a balance between the two. Considering the goals of the task and the characteristics of the graph, selecting the most suitable method requires careful consideration in practical applications.

\section{Formulation of Condensed Graphs}
\label{approachsec}

Here comes the question: how do we formulate each component of the condensed graph $\mathcal{G}_c$?
Since the condensed graphs $\{\mathcal{G}_c,\mathcal{Y}^{\prime}\} = \{\{{\bf A}^{\prime}, {\bf X}^{\prime}\},\mathcal{Y}^{\prime}\}$, therefore, $f(G)$ pertains to formulating these three components. We write: ${\bf A}^{\prime} = f_{\bf A}(G; {\bf W})$, ${\bf X}^{\prime} = f_{\bf X}(G; {\bf W})$, and $\mathcal{Y}^{\prime}=f_{\mathcal{Y}}(G; {\bf W})$ to formulate each.
As we conclude, there are two main classes: the \textbf{Modification} and the \textbf{Synthetic} formulation. 
Specifically:

\subsection{Modification formulation}
Modification approaches encompass actions such as node aggregation and deletion, etc., where the condensed graph is the product of modifying the original graph. 
This category of formulations can be uniformly formalized as aggregating nodes from $\mathcal{G}$ to $\mathcal{G}_c$. Assuming each node $v^{\prime}_i\in V^{\prime}$ is aggregated from $k$ nodes in $\mathcal{G}$, $k\in \mathbb{N}$, then the most common scheme, e.g., \cite{loukas2019graph,deng2020graphzoom,jin2020graph,huang2021scaling,ma2021unsupervised,kumar2023featured,dickens2023graph,gao2023graph} did, was:
\begin{equation}
\begin{aligned}
    f_{\bf A}(\mathcal{G}; {\bf P}) &= {\bf P^{T}AP} \text{ },\text{ } f_{\bf X}(\mathcal{G}; {\bf P}) = {\bf P^{+}X},\\
    f_{\bf \mathcal{Y}}(\mathcal{G}; {\bf P}) &= \arg \max {\bf P^{+}}{[\mathcal{Y}]}
\end{aligned}
\end{equation}
$P \in \mathbb{R}^{N\times N^{\prime}}$ is defined as a projection matrix,
indicating that nodes $\mathcal{V}_{(i)}$ in $\mathcal{G}$ were aggregated to a new node $v^{\prime}_i$ in $\mathcal{G}_c$:
\begin{equation}
\label{projectionmat}
{\bf P}_{i, j}=
    \begin{cases}
        1 & \text { if } v^{\prime}_j \in \mathcal{V}_{(i)} \\
        0 & \text { otherwise }
    \end{cases}
\end{equation}

In a general definition, each row of ${\bf P}$ may contain an uncertain number of nonzero entries, ranging from none (the node is considered dropped, e.g., \cite{luo2021learning}) to one (the node is aggregated once) and even multiple (communities have overlapping issues). No paper has yet delved into the discussion of community overlapping in this field, however, this scenario can also be included within our formulation.

\subsection{Synthetic formulation}
Synthetic approaches, on the other hand, take the condensed graphs as parameters and directly optimize them by minimizing specific objective functions. We further divide this formulation into three strategies: Predefined, Joint Optimization, and Sequential Optimization. Specifically:
\paragraph{Predefined.}
This kind of strategy is undoubtedly the most straightforward yet most tricky one. Two popular strategies are used in the literature: predefine ${\bf A}^{\prime}={\bf I}$ (${\bf I}$ is the identity matrix) in \cite{zheng2023structure,liu2023cat} and predefine $\mathcal{Y}^{\prime}=Sample(\mathcal{Y})$. The former can be interpreted as the goal of graph condensation being solely to learn the prototype embeddings for each class, at which point the topology information has already been integrated and is no longer necessary.
The latter can be explained as achieving the same label distribution between the graph before and after condensation by employing a uniform sampling of labels.
The tricky initialization of parameters to optimize is also included as Predefined strategies, which will not be discussed further here.
\paragraph{Joint Optimization.}

Methods in this category, e.g., \cite{jin2022condensing,xu2023kernel,liu2023graph}, are the most simple yet the most challenging ones, where the condensed graph (topology ${\bf A}^{\prime}$ and node features ${\bf X}^{\prime}$) is considered as parameters for the optimization objective, and the node labels $\mathcal{Y}^{\prime}$ were often predefined by sampling the original labels $\mathcal{Y}$. In conclusion, the formulations are given by: 
\begin{equation}
\begin{aligned}
    f_{\bf A}(; {\bf {A}^{\prime}}) &= {\bf A}^{\prime} \text{ },\text{ } f_{\bf X}(; {\bf {X}^{\prime}})  = {\bf X}^{\prime} \text{ },\text{ }
    f_{\mathcal{Y}}({\mathcal{Y}}; )\subseteq {\mathcal{Y}}
\end{aligned}
\end{equation}

In this scenario, ${\bf W} = \{{\bf A^{\prime}}, {\bf X^{\prime}}\}$ are parameters to be optimized.
So far in the literature, they invariably need to predefine the labels for the condensed graph, e.g., uniform sampling to keep label distribution unchanged. Therefore, this kind of strategy can be perceived as generating dual features for each class: node features and their topological connection.

\paragraph{Sequential Optimization.}

The existence of this strategy is typically regarded as a compromise in the challenge of joint optimization: if the complete condensed graph, encompassing both matrix ${\bf A}^{\prime}$ and vector ${\bf X}^{\prime}$, is regarded as optimization parameters, the dimensionality of the parameter space escalates significantly, introducing challenges to the convergence of optimization objective. Therefore, optimizing part of the condensed graph first, and constructing the rest parts to complete the condensation can be an efficient solution. Specifically, \cite{jin2021graph,liu2022graph,wang2023fast,gao2023multiple,yang2023does} first generate the node embedding via objective optimization, and construct the topology by MLPs according to the generated node features (\cite{gao2023multiple,yang2023does} further introduce original topology ${\bf A}$ to help generate the condensed graph topology ${\bf A}^{\prime}$); and \cite{huang2021scaling,dickens2023graph} modify the condensed graph first, and determine the node label by the majority of aggregated original nodes. The formulation is given by: 
\begin{equation}
\begin{aligned}
    f_{\bf A}({\bf {X}^{\prime}}; {\bf \omega}) &= g({\bf {X}^{\prime}}; {\bf \omega}) \text{ },\text{ } f_{\bf X}(; {\bf {X}^{\prime}})  = {\bf {X}^{\prime}} \text{ },\text{ }
    f_{\mathcal{Y}}({\mathcal{Y}}; )\subseteq {\mathcal{Y}}
\end{aligned}
\end{equation}
$g()$ can be MLPs, etc., and in this case, ${\bf W} = \{{\bf X^{\prime}}, {\bf \omega}\}$.

Among the existing literature, as the optimization of ${\bf X}^{\prime}$ also needs predefined labels, this formulation can be seen as generating prototypes for each class first, and subsequently predicting their relationships (i.e., topology); or aggregate hypernodes first, and determining their labels; or construct topology first, optimizing the node feature and labels as \cite{pan2023fedgkd} did, hence possessing greater interpretability.

\begin{table}[t]     
\footnotesize
\centering                
\renewcommand\arraystretch{1.2}
\caption{Taxonomy of surveyed methods}  
\vspace{-0.2cm}      
\renewcommand{\arraystretch}{1.25}
\begin{tabular}{cccc}

\toprule
\multirow{1}*{\diagbox{}{}}&\multirow{1}*{Method}& \multirow{1}*{$\mathcal{O}$} & \multicolumn{1}{c}{$f$}    \\ \midrule
\multirow{8}*{\rotatebox{90}{Graph Guided}} & GC        $^1$    &   Spectral Properties        & Modification        \\ 
                                            &ReduceG $^{2}$    &   Spectral Properties         & Modification        \\
                                            & SCAL      $^3$    &   Spectral Properties        & Modification        \\ 
                                            & GraphZoom $^4$    &   KNN Aggregation            & Modification        \\ 
                                            & SC        $^{5}$    &   KNN Aggregation          & Modification        \\ 
                                            & FGC       $^6$    &   Graph Statistics           & Modification        \\ 
                                            & CaT       $^{7}$ &   Graph Similarity            & Synthetic        \\  
                                            & OTC       $^{8}$    &   Ranking + Reconstruct    & M+S      \\   \midrule
\multirow{11}*{\rotatebox{90}{Model Guided}} &ConvMatch  $^{9}$    &   Embedding Similarity    & Modification        \\     
                                            & GCDM      $^{10}$ &   Embedding Similarity       & Synthetic        \\   
                                            & KiDD      $^{11}$    &   Loss Matching           & Synthetic        \\ 
                                            & FedGKD    $^{12}$ &     Loss Matching            & Synthetic        \\ 
                                            & GC-SNTK   $^{13}$ &   Loss Matching              & Synthetic        \\ 
                                            & SFGC      $^{14}$ &   Gradient Matching          & Synthetic        \\ 
                                            & DosCond   $^{15}$    &   Gradient Matching       & Synthetic        \\ 
                                            & GCond     $^{16}$ &   Gradient Matching          & Synthetic        \\ 
                                            & GroC     $^{17}$  &   Gradient Matching          & Synthetic        \\       
                                            & HCDC     $^{18}$  &   Gradient Matching          & Synthetic        \\      
                                            &MSGC       $^{19}$ &   Gradient Matching          & Synthetic        \\ \midrule        
\multirow{3}*{\rotatebox{90}{Hybrid}}       & Mcond     $^{20}$ &   Reconstruct + Gradient     & M+S        \\      
                                            & SGDD      $^{21}$ &   Spectral + Gradient        & Synthetic        \\
                                            & GCEM      $^{22}$ &   Spectral + Logits          & Synthetic        \\  \bottomrule
\multicolumn{4}{p{0.45\textwidth}}{\multirowcell{1}[0pt][l]{Ps. 'M', 'S' stands for Modification and Synthetic respectively. \\ 
                                                    Refs: $^{1}$\cite{loukas2019graph}, $^{2}$\cite{bravo2019unifying},\\
                                                    $^{3}$\cite{huang2021scaling}, $^{4}$\cite{deng2020graphzoom}, $^{5}$\cite{jin2020graph},\\
                                                    $^{6}$\cite{kumar2023featured}, $^{7}$\cite{liu2023cat}, $^{8}$\cite{huang2021scaling},\\
                                                  $^{9}$\cite{dickens2023graph}, $^{10}$\cite{liu2022graph}, $^{11}$\cite{xu2023kernel},\\
                                                  $^{12}$\cite{pan2023fedgkd},$^{13}$\cite{wang2023fast}, $^{14}$\cite{zheng2023structure},\\
                                                  $^{15}$\cite{jin2022condensing},  $^{16}$\cite{jin2021graph}, $^{17}$\cite{li2023attend}, \\
                                                  $^{18}$\cite{ding2022faster},$^{19}$\cite{gao2023multiple},$^{20}$\cite{gao2023graph},\\
                                                  $^{21}$\cite{yang2023does},$^{22}$\cite{liu2023graph}
                                                  }}\\ 
\end{tabular}

\label{taxonomy tabel}               
\vspace{2cm}
\end{table}

\subsection{Comparison of formulations}
Each of the formulations mentioned has its distinct method (or not being invented yet but can be done) for generating ${\bf A}^{\prime}$, ${\bf X}^{\prime}$, $\mathcal{Y}^{\prime}$ separately, despite that the Sequential Optimization Formulation must rely on the intermediate results of the other formulations. As we conclude, the \textbf{Modification} formulations exhibit the strongest computational efficiency and interpretability, but their applicability is limited, as each graph awaiting condensation requires the recalibration of the projection matrix. The \textbf{Synthetic by joint optimization} formulation is the simplest, defining the objective and optimizing directly, yet it is also the most challenging, the parameter search space may be so big that convergence is difficult. 
To address this issue, the \textbf{Synthetic by Sequential Optimization} is a 2-step scheme, incorporating the advantages of both easy-to-implement and objective-oriented. The \textbf{Synthetic by Predefined} formulation yields intuitive results and can be effective in carefully designed scenarios. 
The \cref{taxonomy tabel} presents the objective-based category of methods included in this survey, and strategies on how to formulate the condensed graphs.

\subsection{Strategies: $\mathcal{S} \rightarrow \mathcal{S}_c$}
After introducing the condensation of a single graph, we will now explore the strategies for condensing multiple graphs. Currently, there are only a few methods addressing this application, so we will only provide a concise overview:

\textbf{One-by-One strategy:}
If the number of graphs remains unchanged, i.e., every single graph is condensed independently, e.g., \cite{jin2020graph}, we categorize this scheme as the one-by-one strategy. Any method that can condense a single graph can be modified to condense multiple graphs by adopting this strategy.
\textbf{Joint Optimization strategy:}
Similar to the single-graph joint optimization, this strategy combines multiple condensed graphs as parameters of the optimization objective, e.g., {\cite{jin2022condensing}}. By naming the number of graphs in a graph dataset to be $1$, this category of methods would essentially degenerate into a single-graph optimization strategy.
\textbf{Selecting strategy:}
The core of this strategy is to rank each single graph by score functions, and select the top-ranked ones, e.g., \cite{wang2021curgraph,xu2023better}. Given that these methods have limited relevance to our survey, we will not delve further.

\section{Dataset and Evaluation}
\label{datasetandeval}

\subsection{Dataset Statistics}
We systematically organize and summarize the datasets employed in the discussed methods, categorizing them into two primary types: datasets featuring a single large graph and those comprising multiple graphs. The former is typically utilized for tasks such as node classification and edge prediction, while the latter is employed in graph classification. We present key attributes of the datasets, encompassing details such as the number of nodes, number of edges, features and classes, and the graph type (e.g., social network or molecular network) for datasets with a single large graph. Additionally, for datasets containing multiple subgraphs, we provide organization based on the number of subgraphs, average number of nodes, average number of edges, number of labels, and the type of graph. The detailed statistic of the the datasets can be found in our online resources \footnote{ https://github.com/liangliang6v6/GraphCondensation}.

\subsection{Evaluation Metrics}

GC aims to create a significantly smaller graph dataset while preserving sufficient information, thus it is crucial to evaluate how much this information is retained. 
The evaluation of GC methods is challenging compared to the straightforward performance evaluation of traditional GNNs, mainly due to their involvement in multiple aspects.
From a holistic perspective, we summarize the evaluation of the entire GC process into two aspects: effectiveness and efficiency. 
Effectiveness evaluates how well the GC retains the original information, while efficiency
includes both the condensation process and downstream task efficiency. Details can be found below:

\subsubsection{Metrics Over Effectiveness}

From the perspective of input and output, GC methods take the original graphs as input and the condensed graphs as output. 
To verify that the condensed graphs are informative, the effectiveness of GC is evaluated through the following different aspects:
(1) The \textbf{similarity} between the original and condensed graphs is assessed in domains such as spectral and spacial characteristics. 
(2) The \textbf{performance} of condensed graphs in downstream tasks, while closely mirrors evaluations in traditional GNNs, a comparable performance can be considered as a success preservation of valuable information for downstream tasks.
(3) \textbf{Properties} of the condensed graphs alone, e.g.,
applicability which involves integrating GC as a component within an existing system, with evaluation metrics aligning with target systems like Graph Embedding and Graph Continual Learning;
The capabilities such as fairness, generalizability, etc. Specifically:

\paragraph{Similarity with Original Graphs.}
Evaluation of how well condensed graphs replicate the spectral and spacial characteristics of the original graphs involves metrics such as Proportion of Low-Frequency Nodes, Mean of High-Frequency Areas, and various spectral domain metrics. The dissimilarity between recovered graph partitions and ground-truth structures is quantified using Normalized Mutual Information (NMI). Error metrics, such as Relative Eigenerror, Reconstruction Error, Dirichlet Energy, and Hyperbolic Error, also provide insights into the similarity evaluation.

\paragraph{Performance of Mining on Condensed Graphs.}
The evaluation of GC methods entails a diverse set of metrics tailored to specific downstream tasks. In node and graph classification, widely adopted measures include Accuracy (ACC), with the Mean Classification ACC and Standard Deviation providing a more robust assessment. Furthermore, widely adopted performance evaluation metrics, such as the area under the ROC curve (AUROC), area under the precision-recall curve (AUPRC), and F1-score, find extensive application across diverse downstream tasks. For link prediction task, common metrics include Hits@50 and Mean Reciprocal Rank (MRR).

\paragraph{Independent Graphs Property Evaluation.}
The utilization of GC methods in various systems entails the consideration of diverse evaluation metrics. For bias measurement in GC \cite{feng2023fair,mao2023gcare}, fairness metrics, including Demographic Parity (also known as Statistical Parity) and Equal Opportunity, are employed. Considering the fairness and downstream task performance trade-off, they utilize the Pareto frontier Ishizaka and Nemery. The GC methods involved with models' guidance introduce the concept of generalizability, signifying that the condensed dataset can achieve comparable performance across GNNs. Addtionally, guided GNNs may differ from downstream GNNs, demonstrating cross-architecture generalizability.  

\subsubsection{Metrics Over Efficiency}

The fundamental motivation of GC is to facilitate graph mining tasks on large-scale original graphs with efficiency. Consequently, it is imperative to evaluate the computational resources saved by GC in the mining of condensed graphs.
Meanwhile, although GC is an one-time effort, the process of GC itself should not take too much resources.

Within the condensation algorithm, most methods, e.g. \cite{jin2021graph,jin2022condensing}, analyze the \textbf{computational complexity}, primarily emphasizing time complexity and, to a lesser extent, space complexity.
With respect to various datasets and condensation ratios (often expressed as a percentage), these approach entails directly measuring the time required for generating the condensed dataset, commonly referred to as \textbf{condensation time}.
For a more in-depth analysis of this term, certain methods further subdivide the time, e.g., GCEM\cite{liu2023graph} proposes additional metrics: time spent on pre-processing stage; KiDD\cite{xu2023kernel} measures the time dedicated to forward and reverse gradient propagation based on varying batch sizes, etc. 
In addition, considering that the large size of original graph data can cause the Out of Memory (OOM) problems, some methods also measure the \textbf{memory usage} in this process.

The efficiency of training a new GNN in downstream tasks using condensed graphs is undoubtedly superior to that of original graphs, where the time and space saved by GC are empirically proportional to the condensation ratios. 
However, there are still some research \cite{dickens2023graph,wang2023fast,gao2023graph} specifically focused on evaluating downstream task training time and memory usage. 
Notably, not all methods explicitly address the efficiency of GC, but the absence of it does not imply its impossibility to conduct such \cite{jin2020graph,huang2021scaling}. Considering both efficiency and effectiveness, we will introduces a tradeoff consideration
in the following chapters.

\section{Limitations and Challenges}
\label{discuss}

\paragraph{Performance Gap Issue.}

As we analyze, there is an obvious performance gap: the scale of the synthetically condensed graph is considerably smaller than that of the modifying strategies.
For example, the commonly used condensing ratio of synthetic methods is around 1.3\%, while the common condense ratio of modifying the original graph is 30, 50, 70\%. 
Different strategies have their respective advantages, but there might be a significant performance gap when evaluated on the same metric. 
That is to say, the choice of appropriate methods is diverse and depends on the practical needs of application scenarios, which is currently hard to have a unifying strategy.

\paragraph{Efficiency Concern for Applications.}

If the ultimate goal of GC is to train GNNs effectively on scaled datasets, it may be required to ensure that the time and resources invested in GC do not surpass the time saved by training on smaller graphs.
However, since the GC process is a one-time effort, as long as the downstream task continues for a sufficient duration or is repeated frequently, this requirement can still be fulfilled. Thus, the downstream task scenarios must be included as part of the efficiency evaluation scheme.

\paragraph{Comprehensive Effectiveness Metrics.}
Existing methods predominantly evaluate GC effectiveness based on the performance of condensed graphs in downstream tasks. 
However, conventional performance metrics, like classification ACC, may fall short in addressing critical issues such as fairness \cite{mao2023gcare}, robustness against adversarial attacks \cite{li2023attend} etc. 
While few efforts, e.g. \cite{feng2023fair}, have propose constraints focusing on group fairness, future exploration could broaden the scope to include other universal constraint or evaluation metrics for GC.

\paragraph{Unexplored Methodological Capabilities}

The scopes of investigations on generalizability were exploring performances with (1) Various GNN architectures; (2) Model convergence; (3) Extreme condensation ratios and (4) multiple downstream tasks. 
The main idea is to use the model performance on the condensed graph as a metric to assess the quality of the graph condensation. In this context, the performance metric is considered independent of the semantics of the condensed graph, and thus we consider GC methods using this strategy of evaluation to have a lower interpretability.

\section{Future Directions}
\label{future}
While substantial progress has been achieved in GC, we conclude that numerous future explorations persist:

\paragraph{Interpretability of the Condensed Graphs.}

Unlike the natural interpretability of dataset distillation in the field of computer vision \cite{zhao2020dataset}, the output of condensed graphs requires further exploration of interpretability in the real world.
The key identification of GC in our definition is the fact that the nodes or edges in the condensed graph may be \textbf{newly generated}. While these new elements may provide sufficient information for graph mining, their semantic meanings in the real world may be difficult to obtain directly.
Therefore, we believe that enhancing the interpretability of elements in the condensed graph is an important research problem for expanding the application scenarios of GC.

\paragraph{Condensing More Complex Graphs.}

Although GC has been successfully developed in various graphs, most of existing methods have primarily focused on undirected, homogeneous, static graphs.
However, graphs in real-world scenarios are usually more complicated \cite{kazemi2020representation}, such as dynamic graphs(e.g., traffic flow graphs), heterogeneous graphs (e.g., user-item graphs), etc.
Condensation on these complex graphs requires preserving richer information from the original graph, which in turn poses greater challenges.
Due to the complexity and diversity of real-world graph data, more graph types should be considered as well.

\paragraph{Exploring the Correlation between Objectives.}
Under our taxonomy, each single objective of GC can be categorized into two groups: graph-guided and model-guided, by specific information to preserve. These two types of objectives are not inherently conflicting, yet their mutual relation has not been conclusively investigated. For instance, it remains uncertain whether there exists theoretical assurance that the preservation of certain graph properties is sufficient for the retention of GNN performance, or the other way around.

\paragraph{More Graph Formulation.}

While there are already numerous methods for condensed graph formulation, we believe there are still many potential and viable approaches waiting to be explored.
For example, non-uniform sampling of labels during GC might be a viable solution to address label imbalance issues \cite{zhao2021graphsmote}; node features can be predefined as mutually orthogonal one-hot vectors, similar to what was done in \cite{ma2021homophily}, just to generate the topology and thus facilitate relationship learning, etc.

\paragraph{Tradeoff Framework.} 
Within the exploration of applications, we inevitably confront a crucial yet delicate question: 
How to determine the scale of the condensed graph to meet the predefined purpose of GC?
Although some existing methods have recognized the tradeoff between effectiveness and efficiency (as evidenced by their ratio-performance figure), we argues that both effectiveness and efficiency should be comprehensively included in a tradeoff framework. This is crucial to specify the utility of GC and expand its application scope to more practical scenarios. By considering both aspects, we can better understand the benefits and limitations of GC techniques and make informed decisions about their applicability in real-world settings.

\clearpage
\bibliographystyle{named}
\bibliography{ijcai24}

\end{document}